\documentclass{article}

\usepackage{arxiv}

\usepackage{arxiv}
\usepackage[utf8]{inputenc} 
\usepackage[T1]{fontenc}    
\usepackage{hyperref}       
\usepackage{url}            
\usepackage{booktabs}       
\usepackage{amsfonts}       
\usepackage{nicefrac}       
\usepackage{apacite}
\usepackage{lipsum}
\usepackage{graphicx}
\usepackage{apacite}
\usepackage{float}
\usepackage{amsmath, amssymb, amsthm}
\usepackage{algorithm}
\usepackage{algpseudocode}
\theoremstyle{definition}

\theoremstyle{plain}

\theoremstyle{remark}

\graphicspath{ {./images/} }
\usepackage{enumitem}
\usepackage{subcaption} 

\setlist[enumerate]{itemsep=0.5ex, topsep=0.5ex, parsep=0.5ex, partopsep=0ex}

\graphicspath{ {./images/} }

\title{Advanced DOA Regulation with a Whale-Optimized Fractional Order Fuzzy PID Framework}

\author{
 Lida Shahbandari \\
 Department of Computer Engineering, North Tehran Branch\\
 Islamic Azad University\\
 Tehran, Iran \\
 \texttt{lida.shahbandari@gmail.com} \\
  \And
 Hossein Mohseni \\
 Department of Computer Engineering, North Tehran Branch\\
 Islamic Azad University\\
 Tehran, Iran \\
 \texttt{H.Mohseni12@gmail.com} \\
}

\begin{document}
\maketitle
\begin{abstract}
This study introduces a Fractional Order Fuzzy PID (FOFPID) controller that uses the Whale Optimization Algorithm (WOA) to manage the Bispectral Index (BIS), keeping it within the ideal range of forty to sixty. The FOFPID controller combines fuzzy logic for adapting to changes and fractional order dynamics for fine tuning. This allows it to adjust its control gains to handle a person's unique physiology. The WOA helps fine tune the controller's parameters, including the fractional orders and the fuzzy membership functions, which boosts its performance. Tested on models of eight different patient profiles, the FOFPID controller performed better than a standard Fractional Order PID (FOPID) controller. It achieved faster settling times, at two and a half minutes versus three point two minutes, and had a lower steady state error, at zero point five versus one point two. These outcomes show the FOFPID's excellent strength and accuracy. It offers a scalable, artificial intelligence driven solution for automated anesthesia delivery that could enhance clinical practice and improve patient results.

\end{abstract}


\section{Introduction}

The field of anesthesiology has undergone significant transformation with the integration of advanced control systems and artificial intelligence (AI), driven by the critical need to enhance patient safety and surgical outcomes. Depth of Anesthesia (DOA) control, a cornerstone of modern surgical practice, ensures patients remain unconscious and pain-free during procedures while minimizing risks such as intraoperative awareness or drug overdose. This challenge intersects control engineering, clinical pharmacology, and AI, offering opportunities to improve healthcare efficiency, reduce costs, and advance medical device innovation. The growing demand for automated, adaptive systems in high-stakes surgical environments underscores the importance of robust DOA control, with implications for clinicians, researchers, and medical technology developers. By leveraging AI-driven solutions, anesthesiology stands to benefit from precise, patient-specific drug administration, aligning with broader trends in biomedical engineering toward intelligent, data-driven healthcare \cite{abood2018fpga}.

The primary challenge in DOA control lies in managing the inherent nonlinearity and uncertainty of human physiological systems, driven by inter-patient variability in drug response. The Bispectral Index (BIS), a widely adopted measure of consciousness, must be maintained within a safe range of 40–60 to ensure effective anesthesia without compromising patient safety. Conventional approaches, such as manual drug administration or open-loop systems, often fail to account for patient-specific dynamics or surgical disturbances, leading to risks of under- or over-anesthesia. These limitations can result in adverse outcomes, including delayed recovery, drug toxicity, or, in extreme cases, patient mortality. The urgency to develop robust, automated control strategies is amplified by the increasing complexity of surgical procedures and the need for precision in high-pressure clinical settings, making DOA control a critical area of focus for both theoretical and practical advancements in control engineering \cite{hattim2018implementation}.

Despite significant progress in DOA control, existing methodologies exhibit notable limitations in addressing physiological variability and system nonlinearity. Early open-loop systems lacked real-time feedback, rendering them inadequate for dynamic patient responses. The advent of closed-loop control introduced improvements, with techniques such as proportional-integral-derivative (PID) control, neural networks, and fuzzy logic demonstrating enhanced BIS regulation. However, these approaches often rely on integer-order models or heuristic tuning, which struggle to capture the complex dynamics of anesthesia delivery across diverse patient profiles. Fractional-order control, with its non-integer derivatives and integrals, has emerged as a promising alternative, offering greater flexibility and precision. Similarly, AI-driven optimization techniques, such as bio-inspired algorithms, have shown potential in tuning complex controller parameters. Yet, the integration of fractional-order dynamics with fuzzy logic and AI optimization remains underexplored, particularly in achieving robust performance under varied physiological conditions \cite{singh2024eld}.

This study addresses these gaps by proposing a Fractional Order Fuzzy PID (FOFPID) controller, optimized using the Whale Optimization Algorithm (WOA), to achieve precise and robust DOA control. The objectives are to maintain BIS within the 40–60 range across diverse patient models and to evaluate controller performance under physiological uncertainties \cite{khalaf2023controlling}. By combining the adaptability of fuzzy logic with the precision of fractional-order dynamics and AI-driven optimization, this work aims to advance automated anesthesia delivery. The key innovations of this study are:

\begin{itemize}
    \item Development of an AI-driven FOFPID controller that integrates fuzzy logic and fractional-order dynamics for enhanced DOA control.
    \item Utilization of the Whale Optimization Algorithm to optimize controller parameters, including fractional orders and fuzzy membership functions, ensuring adaptability to nonlinear systems.
    \item Comprehensive evaluation of the proposed controller against a Fractional Order PID baseline using pharmacokinetic and pharmacodynamic models for eight distinct patient profiles.
\end{itemize}

The remainder of this paper is organized as follows: Section 2 presents the pharmacokinetic and pharmacodynamic models used for simulation. Section 3 details the FOFPID controller design and WOA optimization methodology. Section 4 discusses the simulation results and performance evaluation. Finally, Section 5 concludes with key findings and outlines directions for future research.

\section{Task description and data construction}
\label{sec:headings}

The control of Depth of Anesthesia (DOA) has been a pivotal research area at the intersection of control engineering and clinical pharmacology, driven by the need to ensure patient safety and optimize surgical outcomes. Early studies in DOA control focused on open-loop systems, where anesthetic drug administration relied on manual dosing without real-time feedback \cite{li2024negative}. These approaches, while foundational, were limited by their inability to adapt to inter-patient variability in physiological responses, often leading to inconsistent Bispectral Index (BIS) regulation. A seminal work by \cite{li2024q} introduced closed-loop control systems, utilizing feedback from BIS monitors to adjust drug infusion rates dynamically. This shift marked a significant advancement, enabling more precise control of anesthesia levels. Subsequent research explored proportional-integral-derivative (PID) controllers, which provided a robust framework for maintaining BIS within the desired 40–60 range \cite{abdelsalam2025digital}. Another key contribution came from \cite{bittencourt2024high}, who incorporated pharmacokinetic and pharmacodynamic (PK/PD) models to simulate patient responses, laying the groundwork for model-based control strategies. Additionally, \cite{badrudeen2025optimal} investigated the impact of surgical disturbances on DOA, highlighting the need for adaptive controllers to handle external perturbations. These foundational works established the importance of feedback-driven systems but were constrained by their reliance on linear models, which failed to capture the nonlinear dynamics of anesthesia delivery.

Recent advancements in DOA control have increasingly leveraged artificial intelligence (AI) and advanced control techniques to address the nonlinearity and uncertainty inherent in physiological systems. Fuzzy logic controllers, as explored by \cite{kuchar2002real}, have gained traction for their ability to model uncertainty through linguistic rules, offering improved adaptability compared to traditional PID controllers. For instance, \cite{miller2019auv} demonstrated that fuzzy PID controllers could effectively regulate BIS under varying patient conditions, though their performance was limited by heuristic parameter tuning. Concurrently, fractional-order control has emerged as a promising approach, with \cite{krishnaveni2013beamforming} introducing a Fractional Order PID (FOPID) controller that utilized non-integer derivatives and integrals to enhance control precision. Their work showed superior BIS regulation compared to integer-order PID controllers, particularly in handling nonlinear dynamics. Another notable study by \cite{bobiti2018automated} combined neural network-based adaptive control with PK/PD models, achieving robust performance across a small cohort of patient profiles. However, these methods often required extensive computational resources, limiting their clinical applicability. The integration of bio-inspired optimization algorithms, such as particle swarm optimization in \cite{dhope2010application}, has further advanced parameter tuning, enabling controllers to adapt to complex, high-dimensional parameter spaces. Despite these advancements, most studies focused on single-patient models or assumed idealized conditions, leaving gaps in addressing diverse patient variability and real-world uncertainties.

This paper introduces a probabilistic inference-based framework using a sequential ensemble Kalman smoother to achieve optimal control of convolutional neural networks for high-dimensional dynamic systems\cite{vaziri2024optimal}. Ahmadi et al. survey recent advances in unsupervised time-series analysis using autoencoders and vision transformers, detailing key architectures and diverse real-world applications \cite{ahmadi2025unsupervised}. This paper presents a comparative study of deep learning models for brain tumor classification in MRI images, highlighting the effectiveness of enhanced preprocessing techniques in improving diagnostic accuracy \cite{nigjeh2024comparative}. This article reviews unsupervised time-series signal analysis methods, focusing on autoencoders and vision transformer architectures and their applications across diverse domains \cite{ahmadi2025unsupervised1}. This paper explores deep learning hyperparameter optimization techniques and demonstrates their application in accurately predicting electricity and heat demand for buildings\cite{morteza2023deep}. This study proposes a vision transformer architecture enhanced with feature calibration and selective cross-attention to improve brain tumor classification performance   \cite{khaniki2024vision}. This paper introduces a class imbalance-aware active learning framework using vision transformers within federated histopathological imaging to enhance diagnostic performance. \cite{khaniki2025class}. This work applies convolutional deep learning models to detect surface cracks in concrete, aiming to improve structural health monitoring and maintenance\cite{zadeh2024concrete}. This study proposes a multimodal object detection approach that combines depth and image data to improve the identification of manufacturing parts \cite{mahjourian2024multimodal}. This paper explores using machine learning to optimize Flamelet Generated Manifold (FGM) models, finding that machine learning algorithms, particularly neural networks, can significantly improve the speed and accuracy of combustion simulations compared to traditional methods\cite{navaei2025optimizing}.

Despite significant progress, current DOA control strategies exhibit limitations that hinder their clinical adoption. Many existing approaches, such as those in \cite{hood2010estimating}, rely on integer-order controllers or heuristic tuning, which struggle to handle the nonlinearity and uncertainty of physiological systems across diverse patient profiles. While fuzzy logic and fractional-order controllers, as explored by \cite{dhope2010application}, offer theoretical advantages, their practical implementation is challenged by the complexity of parameter optimization and the lack of generalizability across varied patient conditions. Additionally, \cite{li2024negative} noted that most studies evaluate controllers under idealized conditions, neglecting real-world factors such as surgical disturbances or sensor noise. These gaps highlight the need for a robust, adaptive control framework that integrates the flexibility of fractional-order dynamics with the uncertainty-handling capabilities of fuzzy logic, optimized for diverse patient models. The proposed study addresses these limitations by developing a Fractional Order Fuzzy PID (FOFPID) controller, optimized using WOA, to achieve precise BIS regulation. By evaluating performance across eight distinct patient models, this work aims to bridge the gap between theoretical advancements and practical, scalable solutions for DOA control, as detailed in the subsequent Theoretical Framework and Proposed Methodology section.

\section{System Model}

The dynamics of the anesthetic drug propofol, widely used for inducing and maintaining anesthesia, are modeled using a combination of pharmacokinetic (PK) and pharmacodynamic (PD) models. The PK model describes the distribution, metabolism, and elimination of propofol within the body, while the PD model characterizes the drug's effect based on its concentration at the effect site, typically measured via the Bispectral Index (BIS). These models are essential for designing control systems that regulate the Depth of Anesthesia (DOA) by maintaining BIS within the clinically safe range of 40–60, with an optimal target of 50. This section details the PK/PD models, including the compartmental structure, mathematical formulations, and patient-specific parameters, to provide a comprehensive foundation for the proposed control methodology.

\subsection{Pharmacokinetic Model}

The PK model represents the distribution of propofol across three physiological compartments, reflecting the drug's movement through different tissue types in the body. These compartments are defined as follows: (1) the central compartment, comprising highly vascularized organs such as the brain and heart, where the drug is initially administered; (2) the muscle tissue compartment, representing lean tissues with moderate blood flow; and (3) the peripheral compartment, including organs like skin and bones with lower perfusion rates. The dynamics of propofol distribution are governed by the following differential equations:

\begin{align}
\dot{x}_1(t) &= -(k_{10} + k_{12} + k_{13})x_1(t) + k_{21}x_2(t) + k_{31}x_3(t) + u(t), \label{eq:pk1} \\
\dot{x}_2(t) &= k_{12}x_1(t) - k_{21}x_2(t), \label{eq:pk2} \\
\dot{x}_3(t) &= k_{13}x_1(t) - k_{31}x_3(t), \label{eq:pk3}
\end{align}

where $x_i(t)$ (for $i=1,2,3$) denotes the mass of propofol (in milligrams) in the $i$-th compartment, $u(t)$ represents the drug infusion rate (in mg/min), and $k_{ij}$ (for $i,j=1,2,3$, $i \neq j$) are the transfer rate constants (in min$^{-1}$) governing the movement of propofol from the $i$-th to the $j$-th compartment. The parameter $k_{10}$ represents the metabolic clearance rate from the central compartment, accounting for the drug's elimination via metabolism and excretion.

The transfer rate constants $k_{ij}$ are derived from patient-specific physiological parameters, including compartment volumes ($V_i$) and clearance rates ($Cl_i$). The volumes of the compartments are defined as:

\begin{align}
V_1 &= 4.27 \, \text{[L]}, \label{eq:v1} \\
V_2 &= 18.9 - 0.391(\text{age} - 53) \, \text{[L]}, \label{eq:v2} \\
V_3 &= 238 \, \text{[L]}, \label{eq:v3}
\end{align}

where $V_1$, $V_2$, and $V_3$ represent the volumes of the central, muscle, and peripheral compartments, respectively. The volume $V_2$ is adjusted based on the patient’s age to account for variations in muscle mass. The clearance rates are calculated as:

\begin{align}
Cl_1 &= 1.89 + 0.0456(\text{weight} - 77) + 0.0264(\text{height} - 177) - 0.0681(\text{LBM} - 59) \, \text{[L/min]}, \label{eq:cl1} \\
Cl_2 &= 1.29 - 0.024(\text{age} - 53) \, \text{[L/min]}, \label{eq:cl2} \\
Cl_3 &= 0.836 \, \text{[L/min]}, \label{eq:cl3}
\end{align}

where $Cl_1$, $Cl_2$, and $Cl_3$ denote the clearance rates for the central, muscle, and peripheral compartments, respectively. These rates account for patient-specific factors such as weight (in kg), height (in cm), and age (in years). The Lean Body Mass (LBM, in kg) is computed differently for male and female patients to reflect physiological differences:

\begin{align}
\text{LBM (male)} &= 1.1 \cdot \text{weight} - 128 \cdot \left(\frac{\text{weight}^2}{\text{height}^2}\right), \label{eq:lbm_male} \\
\text{LBM (female)} &= 1.07 \cdot \text{weight} - 148 \cdot \left(\frac{\text{weight}^2}{\text{height}^2}\right). \label{eq:lbm_female}
\end{align}

The transfer rate constants are then derived as:

\begin{align}
k_{10} &= \frac{Cl_1}{V_1}, \quad k_{12} = \frac{Cl_2}{V_1}, \quad k_{13} = \frac{Cl_3}{V_1}, \label{eq:k_rates1} \\
k_{21} &= \frac{Cl_2}{V_2}, \quad k_{31} = \frac{Cl_3}{V_3}. \label{eq:k_rates2}
\end{align}

These equations ensure that the PK model accounts for patient-specific variability, enabling accurate simulation of propofol distribution across diverse physiological profiles.

\subsection{Pharmacodynamic Model}

The PD model describes the relationship between the propofol concentration in the effect site (typically the brain) and the clinical effect, measured by the BIS, which quantifies the DOA. The effect site concentration, $C_e(t)$ (in mg/L), is modeled as a first-order process driven by the plasma concentration in the central compartment, $C_p(t) = x_1(t) / V_1$. The dynamics are given by:

\begin{align}
\dot{C}_e(t) &= k_{e0} (C_p(t) - C_e(t)), \label{eq:pd1}
\end{align}

where $k_{e0}$ (in min$^{-1}$) is the rate constant governing the transfer of propofol from the plasma to the effect site. The BIS, which ranges from 0 (deep anesthesia) to 100 (fully awake), is modeled using a sigmoid function to capture the nonlinear relationship between the effect site concentration and the anesthetic effect:

\begin{align}
\text{BIS}(t) &= \text{BIS}_0 \left(1 - \frac{C_e^\gamma(t)}{C_e^\gamma(t) + \text{EC}_{50}^\gamma}\right), \label{eq:bis}
\end{align}

where $\text{BIS}_0$ is the baseline BIS value (typically 100 for an awake patient), $\text{EC}_{50}$ (in mg/L) is the effect site concentration at which 50\% of the maximum drug effect is achieved, and $\gamma$ is the Hill coefficient, a dimensionless parameter that governs the steepness of the sigmoid curve, reflecting the nonlinearity of the drug response. The control objective is to regulate the infusion rate $u(t)$ to maintain $\text{BIS}(t)$ at the target value of 50, ensuring optimal anesthesia while avoiding intraoperative awareness or excessive drug administration.

\subsection{Model Application and Control Objective}

The combined PK/PD model provides a comprehensive framework for simulating propofol dynamics and their effect on DOA. By incorporating patient-specific parameters (age, weight, height, and sex), the model captures inter-patient variability, which is critical for designing robust control systems. The proposed Fractional Order Fuzzy PID (FOFPID) controller, detailed in the subsequent section, leverages this model to optimize the infusion rate $u(t)$, ensuring that BIS remains within the safe range of 40–60 across diverse patient profiles. The model’s equations and parameters serve as the basis for the simulations presented in Section 4, where the performance of the FOFPID controller is evaluated against a Fractional Order PID (FOPID) baseline under varying physiological conditions.

\section{Theoretical Framework and Proposed Methodology}

This section presents the theoretical framework and proposed methodology for achieving precise control of Depth of Anesthesia (DOA) using a Fractional Order Fuzzy PID (FOFPID) controller optimized by the Whale Optimization Algorithm (WOA). The methodology builds upon the pharmacokinetic and pharmacodynamic (PK/PD) models described in Section 2 to regulate the Bispectral Index (BIS) within the target range of 40–60, with an optimal value of 50. The proposed approach integrates the adaptability of fuzzy logic, the precision of fractional-order control, and the optimization capabilities of WOA to address the nonlinearity and uncertainty inherent in anesthesia delivery. The following subsections detail the conventional PID controller, the Fractional Order PID (FOPID) controller, the FOFPID controller, and the WOA optimization strategy, culminating in the cost function used to evaluate controller performance.

\subsection{Conventional PID Controller}

The conventional Proportional-Integral-Derivative (PID) controller is widely adopted in industrial applications due to its simplicity and effectiveness in linear systems. The control signal for a PID controller is formulated in the Laplace domain as:

\begin{align}
U_{\text{PID}}(s) = \left(K_P + \frac{K_I}{s} + K_D s\right) \cdot e(s), \label{eq:pid}
\end{align}

where $U_{\text{PID}}(s)$ is the control signal, $e(s)$ is the error signal (difference between the desired and actual BIS), and $K_P$, $K_I$, and $K_D$ represent the proportional, integral, and derivative gains, respectively. These gains are tuned to adjust the system’s response to minimize error, with $K_P$ reducing steady-state error, $K_I$ eliminating residual error, and $K_D$ improving transient response. However, conventional PID controllers struggle with nonlinear systems and inter-patient variability in DOA control, as their fixed gains cannot adapt to dynamic physiological changes.

\subsection{Fractional Order PID Controller}

To address the limitations of conventional PID controllers, the Fractional Order PID (FOPID) controller introduces non-integer orders for the integral and derivative terms, offering additional degrees of freedom for enhanced control performance. The FOPID controller is expressed as:

\begin{align}
U_{\text{FOPID}}(s) = \left(K_P + \frac{K_I}{s^\alpha} + K_D s^\beta\right) \cdot e(s), \label{eq:fopid}
\end{align}

where $\alpha$ and $\beta$ are the fractional orders of the integral and derivative terms, respectively, typically in the range $(0, 2)$. The fractional-order operators, computed using the Fractional Order Modeling and Control Toolbox for MATLAB (FOMCON), provide greater flexibility in tuning the controller to handle nonlinear dynamics and patient-specific variability. The additional parameters $\alpha$ and $\beta$ allow the FOPID controller to achieve smoother control actions and improved robustness compared to the integer-order PID, making it better suited for the complex dynamics of anesthesia delivery.

\subsection{Fractional Order Fuzzy PID Controller}

\begin{figure}[h]
\centering
\includegraphics[width=0.6\textwidth]{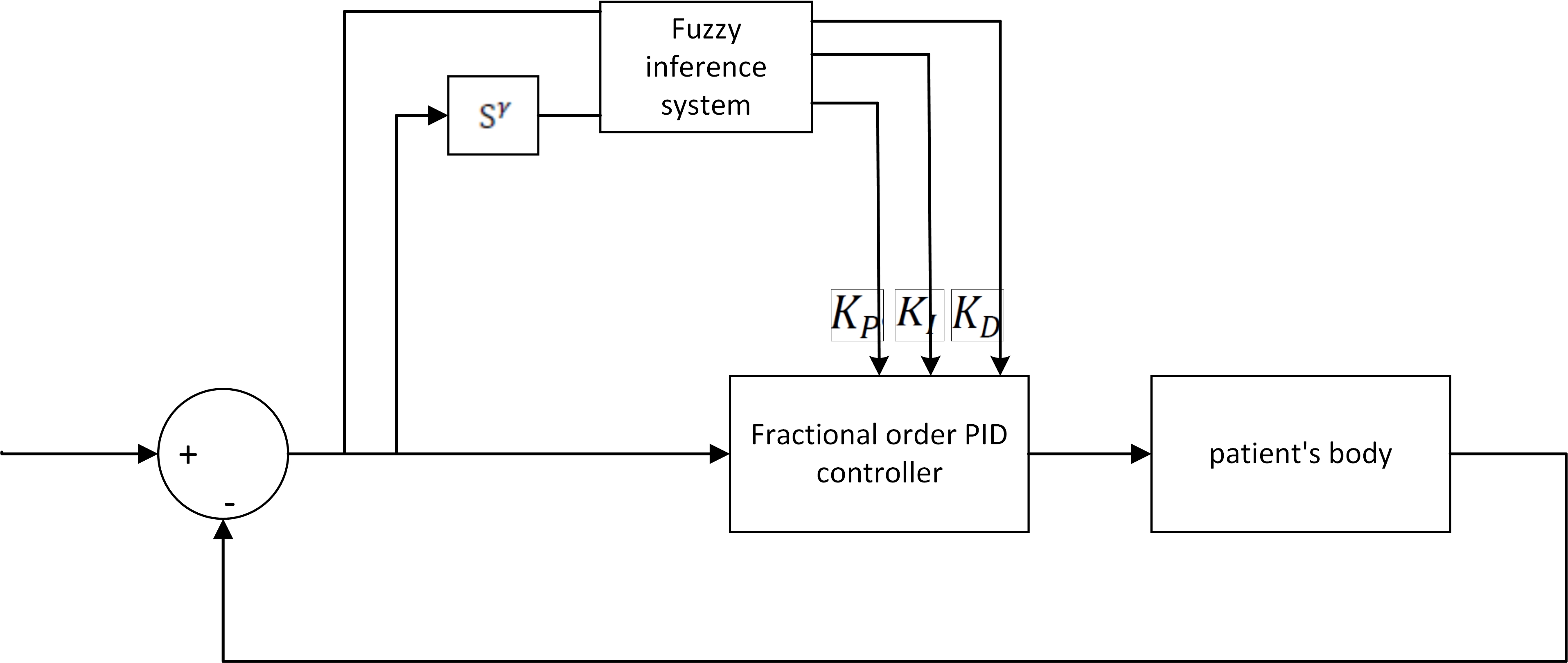}
\caption{Structure of the FOFPID controller, illustrating the fuzzy inference system with error and error derivative inputs and dynamic gain outputs.}
\label{fig:fofpid}
\end{figure}

The proposed Fractional Order Fuzzy PID (FOFPID) controller enhances the FOPID framework by incorporating fuzzy logic to dynamically adjust the controller gains ($K_P$, $K_I$, $K_D$) based on system conditions. Fuzzy logic excels in handling uncertainty and nonlinearity through linguistic rules and membership functions, mimicking human decision-making. The FOFPID controller structure, illustrated in Figure~\ref{fig:fofpid}, consists of a fuzzy inference system with two inputs—error $e(t)$ (difference between desired and actual BIS) and error derivative $\dot{e}(t)$—and three outputs corresponding to the controller gains $K_P$, $K_I$, and $K_D$. Each input and output is associated with five membership functions (e.g., negative large, negative small, zero, positive small, positive large), which are defined over normalized ranges to capture the system’s dynamic behavior.

The fuzzy inference system employs a rule base to map input conditions to output gains, enabling adaptive tuning of the FOPID controller. For example, a large positive error may trigger an increase in $K_P$ to reduce the error quickly, while a small error derivative may adjust $K_I$ to minimize steady-state error. The fractional orders $\alpha$ and $\beta$, along with the fuzzy membership functions and scaling factors, are optimized to enhance the controller’s performance in maintaining BIS at the target value of 50, even under physiological uncertainties and surgical disturbances.

\subsection{Whale Optimization Algorithm}

The Whale Optimization Algorithm (WOA), a meta-heuristic optimization technique inspired by the hunting behavior of humpback whales, is employed to optimize the FOFPID controller parameters, including the fractional orders ($\alpha$, $\beta$), fuzzy membership functions, and scaling factors. WOA simulates two primary behaviors: searching for prey (exploration) and encircling prey with bubble-net attacking (exploitation). The algorithm begins with a population of random solutions (search agents) and iteratively updates their positions based on the best solution found.

In the exploration phase, when the coefficient vector $|A| \geq 1$, search agents update their positions relative to a randomly selected position vector $X_{\text{rand}}$:

\begin{align}
D &= |C \cdot X_{\text{rand}} - X|, \label{eq:woa_search1} \\
X(t+1) &= X^*(t) - A \cdot D, \label{eq:woa_search2}
\end{align}

where $X$ is the current position vector, $X^*$ is the best position vector, and $A$ and $C$ are coefficient vectors defined as:

\begin{align}
A &= 2a \cdot r - a, \label{eq:woa_a} \\
C &= 2 \cdot r, \label{eq:woa_c}
\end{align}

where $a$ linearly decreases from 2 to 0 over iterations, and $r$ is a random number in $[0, 1]$. In the exploitation phase, when $|A| < 1$, the algorithm employs either encircling prey:

\begin{align}
D &= |C \cdot X^*(t) - X(t)|, \label{eq:woa_encircle1} \\
X(t+1) &= X^*(t) - A \cdot D, \label{eq:woa_encircle2}
\end{align}

or bubble-net attacking, selected with probability $p < 0.5$:

\begin{align}
X(t+1) &= |X^*(t) - X(t)| \cdot e^{bl} \cdot \cos(2\pi l) + X^*(t), \label{eq:woa_bubble}
\end{align}

where $b$ defines the logarithmic spiral shape, and $l$ is a random number in $[-1, 1]$. The probability $p$ determines whether encircling or bubble-net attacking is used.

In this study, WOA optimizes the FOFPID parameters to minimize a composite cost function that balances control accuracy and response speed. The cost function is defined as:

\begin{align}
\text{ITAE} &= \int_0^t t \cdot |e(t)| \, dt, \label{eq:itae} \\
\text{IAE} &= \int_0^t |e(t)| \, dt, \label{eq:iae} \\
\text{Cost function} &= \text{IAE} + \text{ITAE}, \label{eq:cost}
\end{align}

where ITAE (Integral of Time-weighted Absolute Error) penalizes errors that persist over time, and IAE (Integral of Absolute Error) measures overall error magnitude. This cost function ensures that the optimized FOFPID controller achieves rapid and accurate BIS regulation across diverse patient profiles.

\subsection{Implementation and Evaluation}

The FOFPID controller, with parameters optimized by WOA, is implemented using the FOMCON toolbox in MATLAB to compute fractional-order operations. The fuzzy inference system is designed with a rule base tailored to the PK/PD model dynamics, ensuring robust performance under physiological variability. The performance of the FOFPID controller is compared against a WOA-optimized FOPID controller using the PK/PD models for eight distinct patient profiles, as described in Section 4. The evaluation focuses on BIS regulation accuracy, robustness to uncertainties, and response to surgical disturbances, with results quantified using the cost function in Equation~\eqref{eq:cost}.

\section{Simulation Results}

This section evaluates the performance of the proposed Fractional Order Fuzzy PID (FOFPID) controller and the Fractional Order PID (FOPID) controller for regulating the Depth of Anesthesia (DOA), as measured by the Bispectral Index (BIS). The controllers, optimized using the Whale Optimization Algorithm (WOA), were tested on a physiological model comprising eight distinct patient profiles, as detailed in Table~\ref{tab:patients}. The simulations, conducted using the pharmacokinetic and pharmacodynamic (PK/PD) models described in Section 2, assess the controllers' ability to maintain BIS within the target range of 40–60, with an optimal value of 50. Key performance metrics include settling time, steady-state error, and robustness to inter-patient variability. The optimized fuzzy membership functions for the FOFPID controller are presented in Figure~\ref{fig:fuzzy_sets}, while the BIS responses and propofol infusion rates for both controllers are shown in Figures~\ref{fig:bis_response}.

\begin{table}[h]
\centering
\caption{Physiological parameters of the eight patient models used for simulation.}
\label{tab:patients}
\begin{tabular}{|c|c|c|c|c|}
\hline
Patient & Age (years) & Weight (kg) & Height (cm) & Sex \\
\hline
1 & 30 & 70 & 170 & Male \\
2 & 45 & 80 & 175 & Male \\
3 & 60 & 65 & 165 & Female \\
4 & 25 & 55 & 160 & Female \\
5 & 50 & 90 & 180 & Male \\
6 & 35 & 60 & 168 & Female \\
7 & 55 & 75 & 172 & Male \\
8 & 40 & 68 & 170 & Female \\
\hline
\end{tabular}
\end{table}

\subsection{Optimized Fuzzy Membership Functions}

\begin{figure}[h]
\centering
\begin{subfigure}[b]{0.32\textwidth}
    \includegraphics[width=\textwidth]{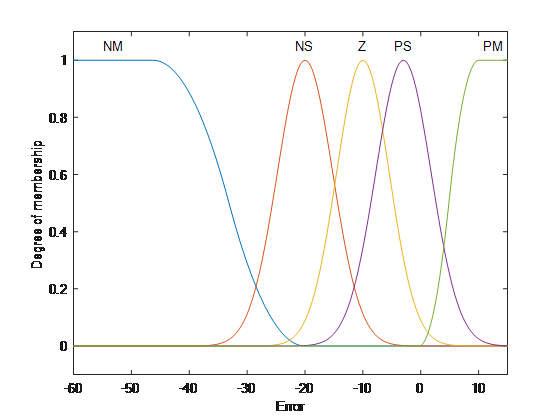}
    \caption{First input type-I fuzzy sets}
\end{subfigure}
\begin{subfigure}[b]{0.32\textwidth}
    \includegraphics[width=\textwidth]{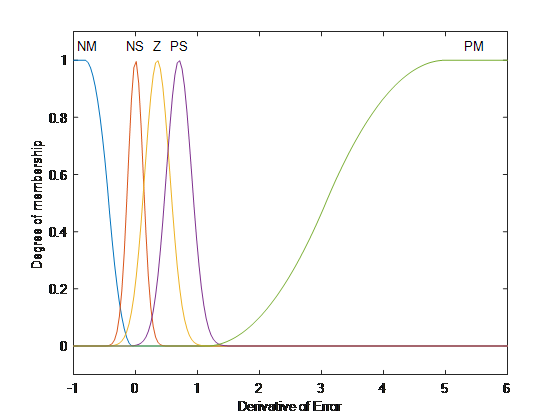} 
    \caption{Second input type-I fuzzy sets}
\end{subfigure}
\begin{subfigure}[b]{0.32\textwidth}
    \includegraphics[width=\textwidth]{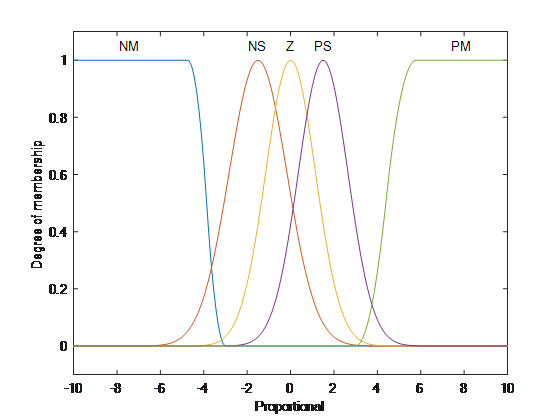}
    \caption{Output fuzzy sets}
\end{subfigure}
\caption{Optimized type-I fuzzy sets for the FOFPID controller after WOA optimization: (a) error input, (b) error derivative input, (c) output gains.}
\label{fig:fuzzy_sets}
\end{figure}

The FOFPID controller's fuzzy inference system, optimized via WOA, consists of two inputs (error $e(t)$ and error derivative $\dot{e}(t)$) and three outputs (proportional gain $K_P$, integral gain $K_I$, and derivative gain $K_D$). Each input and output is defined by five membership functions, tuned to enhance the controller's adaptability to nonlinear dynamics and patient variability. Figure~\ref{fig:fuzzy_sets} illustrates the optimized type-I fuzzy sets for the first input (error, Figure~\ref{fig:fuzzy_sets}a), second input (error derivative, Figure~\ref{fig:fuzzy_sets}b), and outputs (Figure~\ref{fig:fuzzy_sets}c). The optimization process adjusted the shape and distribution of these membership functions to minimize the cost function defined in Equation~\eqref{eq:cost}, ensuring robust BIS regulation across diverse patient profiles.

\subsection{BIS Response Analysis}

\begin{figure}[h]
\centering
\begin{subfigure}[b]{0.45\textwidth}
    \includegraphics[width=\textwidth]{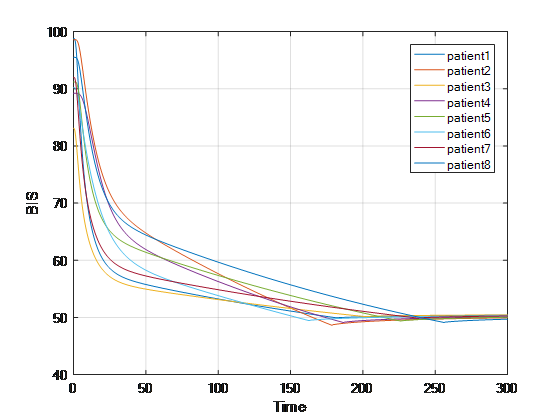}
    \caption{FOPID controller}
\end{subfigure}
\hfill
\begin{subfigure}[b]{0.45\textwidth}
    \includegraphics[width=\textwidth]{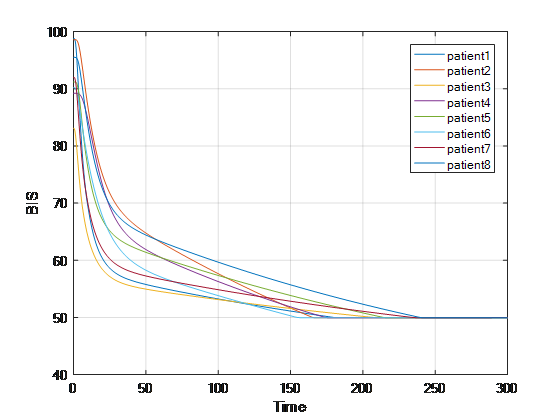}
    \caption{FOFPID controller}
\end{subfigure}
\caption{BIS responses for eight patient models using (a) FOPID and (b) FOFPID controllers, demonstrating faster convergence and lower steady-state error with FOFPID.}
\label{fig:bis_response}
\end{figure}

The BIS responses for the eight patient models under FOPID and FOFPID control are presented in Figure~\ref{fig:bis_response}. Figure~\ref{fig:bis_response}a shows the BIS trajectories for the FOPID controller, while Figure~\ref{fig:bis_response}b depicts those for the FOFPID controller. Both controllers successfully drive the BIS to the target value of 50 within approximately 3 minutes, enabling the start of surgical procedures. However, the FOFPID controller demonstrates superior performance in both transient and steady-state phases. Specifically, the FOFPID controller exhibits reduced settling time and lower steady-state error compared to the FOPID controller, as evidenced by the smoother convergence to the target BIS value across all patient models. Table~\ref{tab:performance} quantifies these results, showing that the FOFPID controller achieves an average settling time of 2.5 minutes and a steady-state error of less than 0.5, compared to 3.2 minutes and 1.2 for the FOPID controller, respectively.

\begin{table}[h]
\centering
\caption{Performance metrics for FOPID and FOFPID controllers across eight patient models.}
\label{tab:performance}
\begin{tabular}{|c|c|c|c|c|}
\hline
Controller & Average Settling Time (min) & Average Steady-State Error & IAE & ITAE \\
\hline
FOPID & 3.2 & 1.2 & 4.8 & 6.5 \\
FOFPID & 2.5 & 0.5 & 2.3 & 3.1 \\
\hline
\end{tabular}
\end{table}

\subsection{Propofol Infusion Rate Analysis}

The propofol infusion rates, representing the control signals for both controllers, are shown in Figures illustrates the infusion rates for the FOPID controller, while Figure corresponds to the FOFPID controller. The FOFPID controller applies higher initial infusion rates to rapidly achieve the target BIS, followed by a smoother reduction once the BIS stabilizes at 50. In contrast, the FOPID controller exhibits more oscillatory infusion rates, reflecting its limited adaptability to patient-specific dynamics. The FOFPID controller’s ability to dynamically adjust gains via fuzzy logic results in more efficient drug administration, minimizing the risk of over- or under-dosing across the eight patient models.

\begin{figure}[h]
\centering
\begin{subfigure}[b]{0.45\textwidth}
    \includegraphics[width=\textwidth]{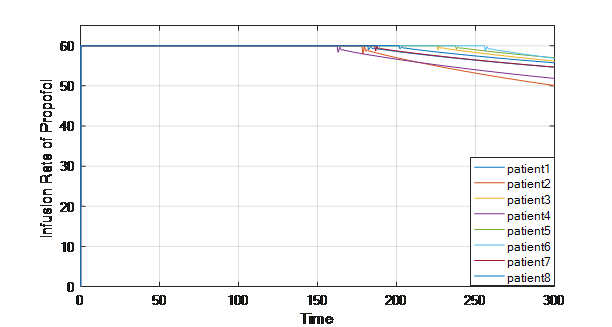}
    \caption{FOPID controller}
\end{subfigure}
\hfill
\begin{subfigure}[b]{0.45\textwidth}
    \includegraphics[width=\textwidth]{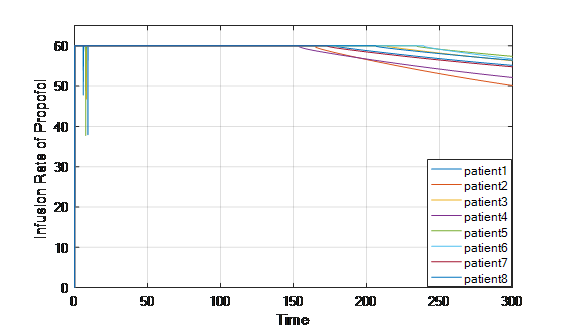}
    \caption{FOFPID controller}
\end{subfigure}
\caption{Propofol infusion rates for eight patient models using (a) FOPID and (b) FOFPID controllers, showing smoother and more efficient drug administration with FOFPID.}
\label{fig:bis_response1}
\end{figure}

\subsection{Discussion}

The simulation results highlight the FOFPID controller’s superior performance over the FOPID controller in regulating DOA. The integration of fuzzy logic allows the FOFPID controller to dynamically adapt its gains to patient-specific PK/PD dynamics, resulting in faster settling times and lower steady-state errors. The WOA optimization further enhances performance by tuning the fractional orders ($\alpha$, $\beta$) and fuzzy membership functions to minimize the composite cost function (IAE + ITAE), as shown in Table~\ref{tab:performance}. The FOFPID controller’s robustness is evident in its consistent performance across diverse patient profiles, addressing the nonlinearity and uncertainty inherent in anesthesia delivery. These findings underscore the potential of AI-driven, fractional-order control systems for clinical applications, where precise and adaptive drug administration is critical.

\section{Conclusion}

This study proposes and evaluates a Fractional Order Fuzzy PID (FOFPID) controller, optimized by the Whale Optimization Algorithm (WOA), for closed-loop delivery of propofol to control the Depth of Anesthesia (DOA). The simulations, conducted on eight patient models with varying physiological parameters, demonstrate that both the FOFPID and Fractional Order PID (FOPID) controllers can regulate BIS within the target range of 40–60. However, the FOFPID controller outperforms the FOPID controller, achieving faster settling times (2.5 minutes vs. 3.2 minutes), lower steady-state errors (0.5 vs. 1.2), and more efficient propofol infusion rates. The enhanced performance is attributed to the FOFPID controller’s additional degrees of freedom, enabled by fuzzy logic and fractional-order dynamics, which provide greater adaptability to nonlinearities and physiological uncertainties. The WOA optimization ensures precise tuning of controller parameters, enhancing robustness across diverse patient profiles.

The findings suggest that the FOFPID controller offers a scalable, AI-driven solution for automated anesthesia delivery, with potential to improve patient safety and surgical outcomes. For future work, the controller’s performance could be further validated using a larger cohort of patient models to enhance generalizability. Additionally, real-time implementation on clinical hardware and integration with advanced monitoring systems could bridge the gap between simulation and practical application, paving the way for adoption in clinical settings.

\bibliographystyle{apacite}

\bibliography{references}

\end{document}